# On Encoding Temporal Evolution for Real-time Action Prediction

Fahimeh Rezazadegan[1,2], Sareh Shirazi[1,2], Mahsa Baktashmotlagh[1], Larry S. Davis[3]
[1]Queensland University of Technology, [2]Australian Centre for Robotic Vision, [3]University of Maryland
fahimeh.rezazadegan@hdr.qut.edu.au

## Abstract

Anticipating future actions is a key component of intelligence, specifically when it applies to real-time systems, such as robots or autonomous cars. While recent works have addressed prediction of raw RGB pixel values, we focus on anticipating the motion evolution in future video frames. To this end, we construct dynamic images (DIs) by summarising moving pixels through a sequence of future frames. We train a convolutional LSTMs to predict the next DIs based on an unsupervised learning process, and then recognise the activity associated with the predicted DI. We demonstrate the effectiveness of our approach on 3 benchmark action datasets showing that despite running on videos with complex activities, our approach is able to anticipate the next human action with high accuracy and obtain better results than the state-of-the-art methods.

## 1 Introduction

While recent computer vision systems can recognize objects, scenes and actions with reasonable accuracy (Russakovsky et al. 2014; Zhou et al. 2014; Wang et al. 2017), anticipating future activities still remains as an open problem which requires modeling human activities, their temporal evolution, and recognizing the relationships between objects and human in the scene.

Predicting human action has a variety of applications from human-robot collaboration and autonomous robot navigation to exploring abnormal situations in surveillance videos and activity-aware service algorithms for personal or health care purposes. As an example, in autonomous healthcare services, consider an agent- monitoring a patient's activities, trying to predict if the patient is losing her/his balance. If the agent is capable of predicting the next action, it could determine whether s/he might fall and take an action to attempt to prevent it. Another example of activity prediction in the robotics area is intelligent human-robot collaboration. For robots and humans to be cooperative partners who share tasks naturally and intuitively, it is essential that the robot understands actions of the human and anticipates the human's needs e.g. the need for tools and parts in assembly tasks.

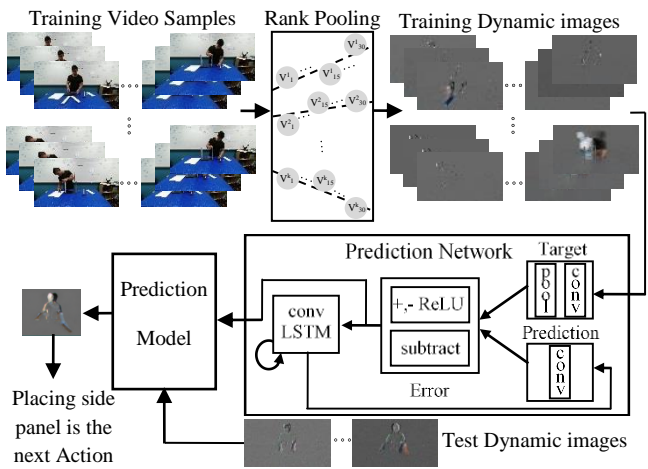

Figure 1: The overview of our human action prediction system.

Prior research focused on future frame prediction with the goal of predicting as many future frames as possible (Lotter, Kreiman and Cox 2016), or predicting trajectories of people, either holistic or of their parts, and then applying activity recognition to those predicted trajectories (Bennewitz et al. 2005; Kitani et al. 2012).

To overcome the limitation of the current approaches on predicting single frame or trajectory of people resulting in inaccurate prediction, our framework introduces an LSTM model to predict semantics of future human activity i.e. human pose and objects' shapes by constructing dynamic images (DIs[1]) in its input layer (Bilen et al. 2016).

Inspired by the previous work in next-frame video prediction (Lotter, Kreiman and Cox 2016), our LSTM model continually predicts future images, and leverages deep recurrent convolutional networks with both bottom-up and top-down connections. Our framework capitalises on the temporal structure of unlabeled videos to learn to anticipate both actions and objects in the future. Our proposed framework is experimentally validated on several benchmark da-tasets of MPII Cooking dataset (Rohrbach et al. 2012), 50 Salads dataset (Stein and McKenna 2013) and IKEA assembly dataset (Eich, Shirazi and Wyeth 2016), which contain complex activities composed of several actions. The main contributions of this work are as follows:

- For the first time in the action prediction literature, we predict a sequence of (up to) 150 frames in a form of dynamic image (DI), performing 5 times longer prediction than the existing works.
- We summarise short video clip in a single image by constructing DIs which gives the potential of going further in the future. We show that predicting a representation like DIs more directly encodes activity than either future frames or future optical flow fields, while gives us critical cues about the performed action resulting in more accurate and long-term activity prediction compared to existing approaches.
- The DI-based Semantic representation of an action in our approach helps in accurately predicting the action before its occurrence. This is different from the existing works on early action recognition which rely on the information of observing the first part of the action for prediction.

---

[1] A DI is a representation incorporating cues of human poses and object shapes in a sequence of frames, originally introduced as a representation for video activity recognition.

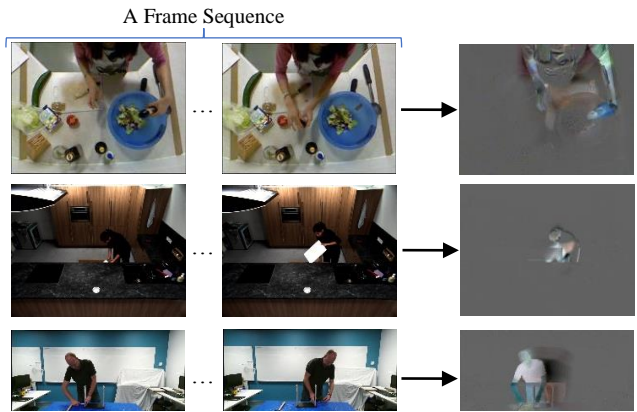

Figure 2: Generating Dynamic images from video frames of 50 Salads dataset, Cooking activities dataset and IKEA dataset, respectively. Here, we show dynamic images generated by using a window size of 30 and a stride of 5.

## 2 Related Work

Action prediction can refer to either early recognition of human activity from a partially observed video containing a single action or, more generally, online prediction of the next activity in a multi task activity video.

There have been a few previous approaches for predicting future actions based on predicting human trajectories in 2D from visual data (Bennewitz et al. 2005; Kitani et al. 2012). For example, Jiang et al. predicted the action movements a person may take in a human-robot interaction scenario using RGB-D sensors and an anticipatory temporal CRF model (Jiang and Saxena 2014). Such works not only enhance the quality of interactions between the human and robot, but also are beneficial to reduce risk assessment expenses (Rezazadegan, Gengb, et al. 2015). However, their accuracy significantly drops for a long anticipation horizon.

A related problem to action prediction is early detection of an action -detect/classify an incoming temporal sequence as early as possible (Ryoo 2011; Lan, Chen and Savarese 2014; Soomro, Idrees and Shah 2016; Sadegh Aliakbarian et al. 2017).

Another group of researchers approach action prediction by leveraging the association between humans and scene elements (Vu et al. 2014), which was shown to be beneficial for action recognition as well (Rezazadegan, Shirazi, et al. 2015; Rezazadegan et al. 2017).

To the best of our knowledge, Li et. Al. is the only work that addressed long-duration activity prediction using sequential pattern mining to incorporate context into prediction (Li and Fu 2016). The temporal predictions called action-lets are based on motion velocity peaks that are difficult to obtain in real-world scenarios.

The first algorithm for video frame prediction inspired by language modeling was presented in Ranzato et al. (Ranzato et al. 2014). Later on, other algorithms were introduced using LSTM, ConvNets, generative adversarial training and predictive coding architectures (Srivastava, Mansimov and Salakhudinov 2015; Mathieu, Couprie and LeCun 2015; Lotter, Kreiman and Cox 2016; Finn, Goodfellow and Levine 2016; Vondrick, Pirsiavash and Torralba 2015).

A convolutional LSTM architecture was employed in some recent papers on future frame prediction (Finn, Goodfellow and Levine 2016; Lotter, Kreiman and Cox 2016; Neverova et al. 2017; Luo et al. 2017). For instance, Finn et al. developed an action-conditioned video prediction model that incorporates appearance information in previous frames with motion predicted by the model and is able to generate frames 10 time steps into the future.

Among all the above-mentioned approaches for future frame prediction, Lotter et al. achieves good results for predicting 5 future time steps at 10 fps, i.e. 0.5 seconds, and Finn et al. predict 10 future time steps, at 10 fps, i.e. 1 second. However long-duration prediction has not been addressed, although it is critical for action prediction, particularly in real-time robotics applications.

Our work targets longer-duration action prediction by summarising a sequence of future frames in a compact representation rather than predicting individual frames. Our extensive experiments on the benchmark action prediction datasets demonstrate the effectiveness of our approach on achieving reasonably accurate predictions to time horizons of 5 seconds (150 frames). We employ the capability of deep networks to predict DIs of future video clips.

## 3 Overview of the Proposed Approach

To make a human action prediction system useful for real-time applications such as robotics, we need to reduce the algorithm latency as much as possible and predict longer in the future. In this work, we predict a compact representation of the sequence of future frames in the forms of DIs to reduce latency in prediction. Our DI-based representation helps in suppressing background information and encoding motion of actors and objects involved in the actions. Being compact (i.e., a few seconds of video are encoded in a single frame) and fast to compute, action prediction from DIs is computationally efficient. Figure 1 shows an overview of our approach to human action prediction.

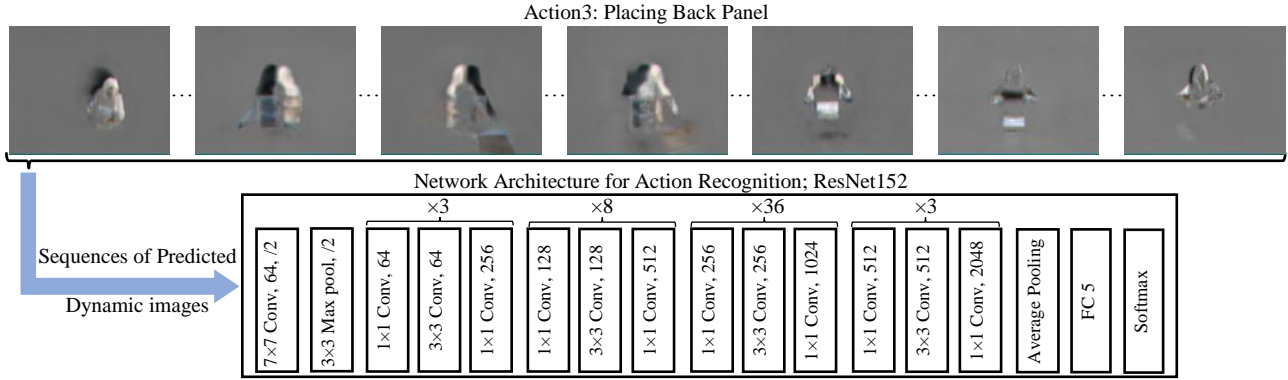

Figure 3: Training a network for action recognition from predicted dynamic images as inputs.

### 3.1 Constructing DIs from Video

A dynamic image (DI) is based on rank pooling and is obtained through the parameters of a ranking machine that encodes the temporal evolution of frames of video. More precisely, DIs focus on those areas of images which contain actors and the objects they interact with, whereas they suppress background pixels and motion patterns which can be considered as noise for action recognition and prediction. As a result, the energy in DIs is concentrated in the salient actors' motion. DIs are constructed by identifying a function $\psi$ that maps a sequence of video frames $(I_1,...,I_T)$ into a video descriptor $d^* = \rho(I_1,...,I_T;\psi)$, which contains enough information to rank all the frames in the video where $\psi(I_t)$ stacks the RGB components of each pixel in image into a large vector. The ranking function, $\rho$, associates to each time t a score $S=<d,V_t>$ where $d$ is a vector of parameters and $V$ is the time average of $\psi$:

$$V_t = \frac{1}{t}\sum_{T=1}^{t}\psi(I_T) \quad (1)$$

Equation 2 reflects the process of learning d and constructing $d^*$ from a sequence of video frames, which is called rank pooling. Note that d is learned so that the scores in later frames are associated with larger scores.

$$d^* = \arg\min(\frac{1}{2}\lambda\|d\|^2 + \frac{2}{T(T-1)}\times\sum_{q>t}\max(0,1-S(q|d)+S(t|d))) \quad (2)$$

### 3.2 Predicting the Next Action

To train a deep model for predicting the next dynamic image of a video, we first construct DIs for 30 frame video clips and use a stride of 5 frames as we pass the one second window through a long video. Then we train a deep model for predicting the next dynamic image of a video. Figure 2 shows examples of 3 datasets and the generated DIs using this pipeline for 1 second clips. The model for predicting dynamic images (DIs) is based on recurrent neural networks and convolutional long short-term memories, extending the architecture of PredNet (Lotter, Kreiman and Cox 2016). The model consists of stacked modules which generate local predictions of the input, and then the logarithmic error between the local predictions and the actual input is flowed to the next layer of the network. Each module of the network contains four layers - an input convolutional layer, $A_l$, a recurrent representation layer, $R_l$, a prediction layer, $\hat{A}_l$, and an error representation, $E_l$ (Eq. 3).

$$\begin{aligned}
A_l^t &= \begin{cases} x_t & l=0 \\ MAXPOOL(RELU(Conv(E_{l-1}^t))) & l>0 \end{cases} \\
\hat{A}_l^t &= RELU(Conv(R_l^t)) \\
E_l^t &= [RELU(A_l^t - \hat{A}_l^t); RELU(\hat{A}_l^t - A_l^t)] \\
R_l^t &= ConvLSTM(E_l^{t-1}, R_l^{t-1}, Upsample(R_{l+1}^t))
\end{aligned} \quad (3)$$

The input layer, prediction layer and error representation layer have the standard shape of deep ConvNets, whereas recurrent representation layers follow the rules of generative deconvolutional networks. Prediction of a DI is based on the last 10 DIs constructed from the previously observed portion of the video. Since each of the DIs summarises a one second video clip, the recurrent network integrates information over 75 observed frames to construct a compact prediction of the next 30.

In principle, one could predict DIs for either longer future time intervals or multiple shorter intervals, although we do not investigate these alternative prediction structures here. However, we consider extrapolating 5 steps further in the future by predicting 5 future DIs. The prediction accuracies for 5 time steps, i.e. 5 seconds into the future, on 3 datasets are presented in Section 5.1, Figure 5.

To predict the next human action, a feature extraction system and a classifier are required to recognize action classes from the predicted DIs. Therefore, this part of the system is based on supervised learning, in contrast to the DI prediction pipeline (Section 3.2) which was unsupervised. Note that typically there are gaps of inactivity between consecutive sub-actions in a video. For example, in one video after the actor completes the "using screw driver sub-action" there is a 3 second gap before the actor initiates the "placing back panel" sub-action. In general, we label the gap with the activity label of the next sub- activity as it is the desired sub-activity we wish to predict. Figure 3 reflects the training process on predicted DIs of IKEA samples and the corresponding action labels.

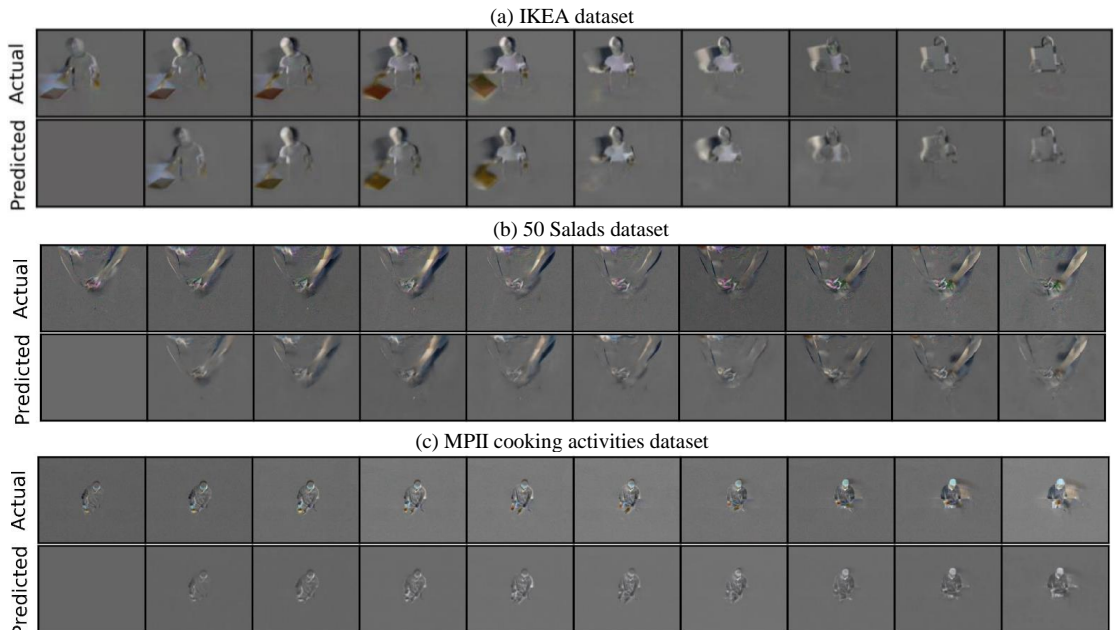

Figure 4: Predicted DIs versus actual ones on (a) IKEA; (b) 50 Salads, (c) MPII cooking datasets; FPS=30, window size=30 and stride=5.

## 4 Experimental Setup

In our experimental evaluations, we apply our method to predict the next DI on 3 public datasets (Rohrbach et al. 2012; Stein and McKenna 2013; Eich, Shirazi and Wyeth 2016). We compare the MSE of predicted DI individually with a simple baseline in which we copy the previous DI. Then, we recognize the action labels of predicted DIs through a ConvNet framework which is the principal objective of our research and compare it to 3 other baselines and two state-of-the-art-models for action prediction. Finally, we conduct an experiment to determine how early we can predict the next action compared to other methods.

### 4.1 Datasets

Since existing methods for action prediction mostly focused on early action recognition, they used datasets containing a single action per video. Differently, we aim to predict the next action by seeing a part of the current one. Therefore, we consider 3 public datasets; each contains multiple actions per video.

The first dataset is MPII Cooking activities dataset which is suitable for evaluating fine-grained action classification (Rohrbach et al. 2012). It contains 44 videos, composed of 65 fine-grained cooking actions such as cut slices, pour, spice and so on, performed in the same kitchen setting. This dataset covers gender, subject and recipe diversity for making 14 different dishes.

The second dataset is 50 Salads dataset, which is composed of 50 sequences of a mixed salad preparation task with two sequences per subject (Stein and McKenna 2013). There are gender and age diversity in videos, and a different task-ordering for each sequence that makes it challenging.

The third dataset is the IKEA dataset which is specifically designed for collaborative robotics challenges, contains 50 videos of different people using different setups for the task of assembling an IKEA drawer (Eich, Shirazi and Wyeth 2016). All actors follow the same instructions, but vary the order of sub-actions, the duration of each sub-action and the locations of assembly tools. We consider 5 action labels in total for the task.

### 4.2 Implementation Details

For the first experiment, we initially extracted video frames of all datasets at 30 fps. We allocated 40 videos as the training set, 2 as the validation set and 2 as the test set for MPII cooking dataset. For the 50 Salads dataset, training set, validation set, and test set contain 46 videos, 2 videos and 2 videos, respectively. For the IKEA dataset, training set contains 41 videos, each has around 1000 frames, while we had 6 videos for testing and 3 videos for validation.

Then, we constructed dynamic images (DIs) following the strategy outlined in Section 3.1 using a window size of 30 and stride of 5. Note that the DIs for clips from different classes are likely to have many common background pixels because the actors tend to be centered in the field of view of the camera. This biases the error signal during training and results in insufficiently accurate prediction of DIs. We found that adding low variance white noise to the inputs during training, and computing the error based on the noisy predictions, yielded a more accurate prediction network. Therefore, sequences of 10 noisy DIs are sampled from datasets, center-cropped and downsampled to 128x160 pixels to be fed into the 4-layer architecture with 3by3 convolutions and layer channel sizes of (3,48,96,192) for the prediction process. For unsupervised training, we used the Adam optimization algorithm using a loss computed only on the lowest layer ($L_0$). We initially set Adam parameters to α=0.001, $β_1$=0.9, $β_2$=0.999, in which the learning rate, α, will be decreased to 0.0001 halfway through training. We also replaced $L_1$ error with logarithmic error in $E_l$ formula, stated in Eq. 3 as it improved the prediction results.

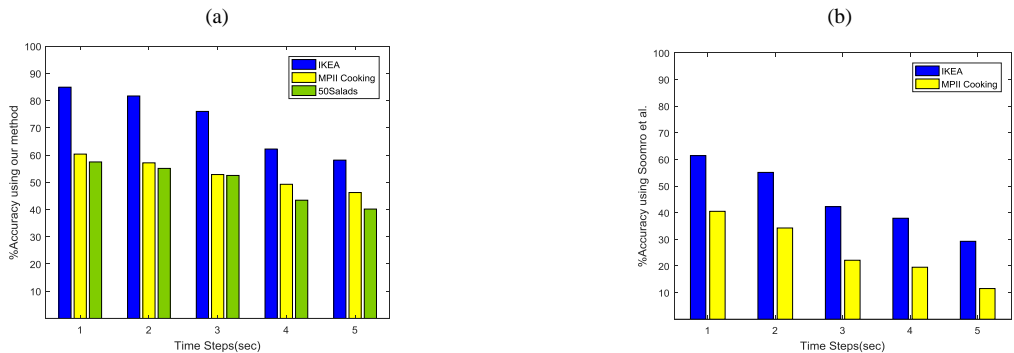

Figure 5: Accuracy of prediction over 5 time steps (5 seconds) into the future using a) ours on 3 datasets; b) Soomro et al. on 2 datasets.

In order to create a model to recognize the action class of predicted DIs for unseen test videos, we then finetune the ConvNet on our predicted DIs from Section 3.2, which are resized to 224×224. Note that we use a model that we finetuned on Sport-1M dataset, for finetuning. This network contains 152 layers including convolutional layers and skip connections ending with a global average pooling layer and a fully connected layer with softmax same as (He et al. 2016), originally proposed for object recognition. The parameters of the network were optimized using stochastic gradient descent with a learning rate of 0.0001, reduced by the factor of 0.5 after every 5k iterations, a momentum of 0.9, a weight decay of 0.0005, and mini-batches of size 16. We present final action prediction results for 3 datasets as the accuracy of recognized actions from predicted DIs.

## 5 Experimental Results

In this section, we first present results of the prediction algorithm for 3 above-mentioned datasets for 1 time step (Figure 4). Then, we extend the predictions up to 5 time steps (Figure 5). Finally, we explore how early we can predict actions for IKEA test videos in average (Figure 6).

Figure 4 shows examples of predicted dynamic images (DIs) versus actual DIs for all 3 datasets. In addition to these qualitative results, to evaluate how accurate the predicted DIs are compared to their ground truth, we calculated Model MSE, and compared with the Prev. MSE in Table 1, to have a fair comparison with the employed method in (Lotter, Kreiman and Cox 2016). Model MSE is the mean squared error between the predicted DI and the actual DI for the next second of video. Prev. MSE is mean squared error between the DI for the last second of observed video and the actual DI of the next second. Table 1, shows about a 50% improvement in Model MSE versus Prev. MSE, for 3 datasets.

To evaluate how accurately we can predict actions from predicted DIs, we report the recognition accuracy of the predicted DIs for unseen test videos of 3 datasets, using our recognition model which we trained based on ResNet152 architecture (He et al. 2016). Table 2 shows the average of prediction accuracies for all actions of the test videos for all datasets, and compares our method to two state-of-the-art methods (Soomro, Idrees and Shah 2016; Sadegh Aliakbarian et al. 2017).

Table 1: Model MSE compared to MSE between previous Dynamic images (DIs), on 3 public datasets.

| Datasets | Model MSE | Prev. MSE |
|---|---|---|
| 50 Salads | **0.003696** | 0.005319 |
| MPII Cooking | **0.002562** | 0.003472 |
| IKEA | **0.001131** | 0.002321 |

For (Soomro, Idrees and Shah 2016) since their approach estimated human pose, we were unable to repeat the experiment for 50 Salads dataset which does not contain the human body. Due to unavailability of the code for (Sadegh Aliakbarian et al. 2017), we report the accuracy only for IKEA dataset which the author has kindly provided us.

Table 2: Our final prediction mean accuracy of one second ahead, on all actions of 3 datasets.

| Datasets | Ours | Soomro et al. | Sadegh et al. |
|---|---|---|---|
| MPII Cooking | **60.36%** | 40.51% | N/A |
| 50 Salads | **57.48%** | N/A | N/A |
| IKEA | **84.94%** | 61.42% | 76.67% |

We reported the accuracy of predicting the action of one future time step, i.e. one second into the future, for all baselines in Table 3. Our method using DIs (the 1st column of Table 3) outperformed all 3 baselines. It verifies the strong ability of our predicting model, simultaneously exploiting the temporal and pose information hidden in the DIs as inputs. There are some incorrect labels for sub-actions that are caused by appearance similarity of sub-actions in datasets.

Table 3: A comparison of mean accuracy of our approach on next action prediction for all actions of test videos in 3 datasets, with 3 baselines: A) Using the previous DI as prediction; B) Labelling the original DIs with the class of next action and training a CNN on these differently labelled DIs for action recognition; C) Training a LSTM to predict the future action.

| Datasets | Ours | A | B | C |
|---|---|---|---|---|
| MPII Cooking | **60.36%** | 48.52% | 28.14% | 32.25% |
| 50 Salads | **57.48%** | 46.82% | 27.5% | 31.14% |
| IKEA | **84.94%** | 75.76% | 41.49% | 46.06% |

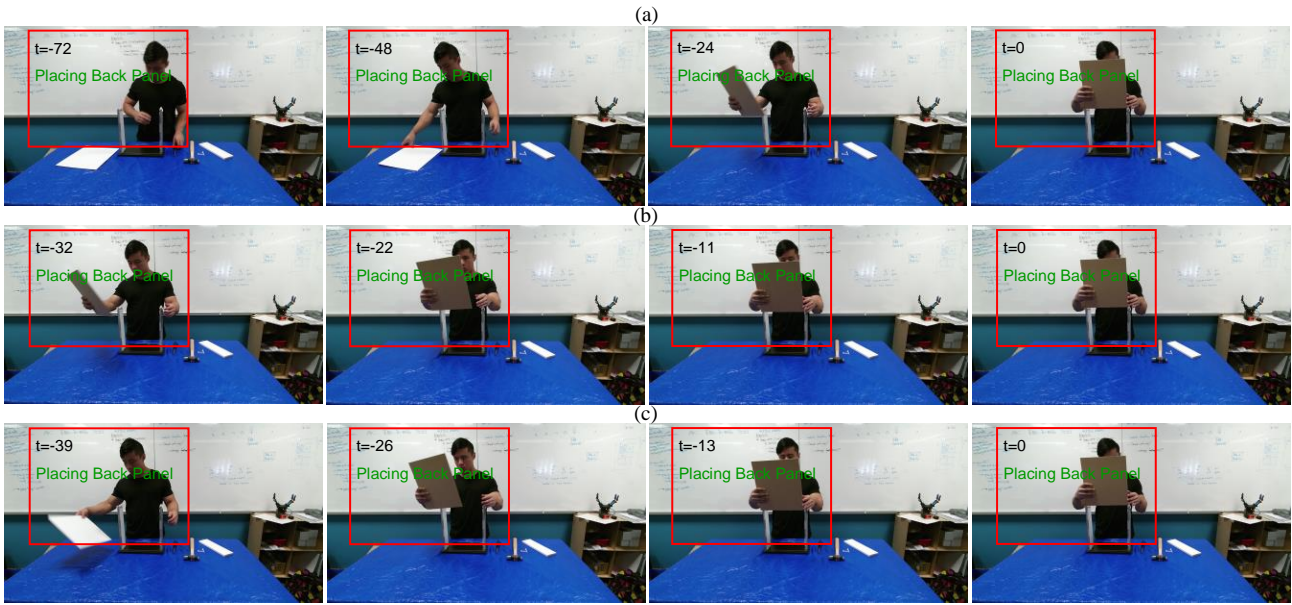

Figure 6: Reflecting how early the next action can be predicted on a sample video, using a) our method; b) Soomro et al.; c) Sadegh et al. For example, the first row shows that our method predicted that "Placing Back Panel" is going to happen, 72 frames before (t=-72) than it occurs (t=0); the prediction was not change from the first prediction point to its occurrence time.

Note that there is a gap between the time that the network predicts the correct label for the next action and the time that the human starts the action (for example, the time when a required tool is picked up for the next action). This is consistent with our final goal - passing tools by a robot before the human needs them.

### 5.1 Further time step prediction

We consider extrapolating 5 steps further in the future by finetuning our prediction network by feeding back predictions as the inputs and recursively iterating. More precisely, we train the model with the loss over 15 time steps, starting from the trained weights. In fact, we feed the original dynamic images (DIs) into the network for the first 10 and then we use the model's predictions for the remaining 5 inputs. This covers predictions for the next 5 time steps (5 seconds). Figure 5 shows a bar graph representing the trend of degrading prediction accuracy compared to the state-of-the-art method (Soomro, Idrees and Shah 2016). Note that since (Soomro, Idrees and Shah 2016) is based on estimating pose and human model, we excluded the 50 Salads dataset in which we do not have the human body, but only hands. The proposed method in (Sadegh Aliakbarian et al. 2017) is not capable of predicting up to 5 seconds and excluded from this experiment. Despite eventual blurriness particularly for the fourth and fifth future predicted DIs, which can be expected due to uncertainty, the fine-tuned model captures some key structures in its extrapolations which is still useful for predicting actions.

### 5.2 Measuring temporal distance

In order to investigate our method's ability in predicting future actions at different stages and various actions, we evaluated the performance in terms of the temporal distance (frames) between the current image and the starting frame of the next action, for IKEA dataset. Table 4 reports the average of these temporal distances (AvgOfTD) on 6 test videos, for all actions placing side panels, placing back panel, placing rear panel, using screw driver and fixing bolts, using our method compared with other methods. Note that we used one time step (1 second ahead) prediction here. We also showed predicted actions using our method compared with two other methods at different temporal stages for a test video sample (run42, action3), in Figure 6.

Table 4. A comparison between AvgOfTDs obtained from our method and other state-of-the-art methods.

| AvgOfTD | Action1 | Action2 | Action3 | Action4 | Action5 |
|---|---|---|---|---|---|
| Ours | 33.6 | **30** | **63.3** | **42** | 30 |
| Soomro et. al | 13.2 | 13.1 | 28.7 | 17.6 | 11.3 |
| Sadegh et. al | **51.5** | 21.8 | 23.5 | 41.8 | **36.3** |

## 6 Conclusion

We proposed an action prediction framework based on a convolutional LSTM model to predict a future DI that is the summary of 30 future RGB frames. We showed the ability of our method to predict up to 5 dynamic images (DIs) whose corresponding actions can be recognized by a ConvNet. We attained 85%, 60% and 57% accuracy for predicting a human action one second before it happens on IKEA, MPII cooking and 50Salads datasets, respectively, which degrades to 58%, 46% and 40% for the fifth time step (5 seconds ahead). Experiments verified the superiority of our model on 3 datasets- for motion encoding and predictive learning compared to 6 baselines and two state-of-the-art methods for action prediction.